# Deep Metric Learning Assisted by Intra-variance in A Semi-supervised View of Learning

Insert Subtitle Here

Liu Pingping†
College of Computer Science and Technology
Jilin University
Changchun China
liupp@jlu.edu.cn

Liu Zetong
College of Computer Science and Technology
Jilin University
Changchun China
ztliu21@mails.jlu.edu.cn

Lang Yijun
College of Computer Science and Technology
Jilin University
Changchun China

## ABSTRACT

Deep metric learning aims to construct an embedding space where samples of the same class are close to each other, while samples of different classes are far away from each other. Most existing deep metric learning methods attempt to maximize the difference of inter-class features. And semantic related information is obtained by increasing the distance between samples of different classes in the embedding space. However, compressing all positive samples together while creating large margins between different classes unconsciously destroys the local structure between similar samples. Ignoring the intra-class variance contained in the local structure between similar samples, the embedding space obtained from training receives lower generalizability over unseen classes, which would lead to the network overfitting the training set and crashing on the test set. To address these considerations, this paper designs a self-supervised generative assisted ranking framework that provides a semi-supervised view of intra-class variance learning scheme for typical supervised deep metric learning. Specifically, this paper performs sample synthesis with different intensities and diversity for samples satisfying certain conditions to simulate the complex transformation of intra-class samples. And an intra-class ranking loss function is designed using the idea of self-supervised learning to constrain the network to maintain the intra-class distribution during the training process to capture the subtle intra-class variance. With this approach, a more realistic embedding space can be obtained in which global and local structures of samples are well preserved, thus enhancing the effectiveness of downstream tasks. Extensive experiments on four benchmarks have shown that this approach surpasses state-of-the-art methods.

## 1 INTORDUCTION

Deep metric learning aims to learn an embedding space where samples of the same class are close and samples of different classes are far from. In this way, deep metric learning captures semantic similarity information between samples and has excellent generalization over unseen classes [1]. With its robust instance representation capabilities, deep metric learning is now widely used in computer vision tasks, including content-based image retrieval [2, 3], medical image processing [4-6], person re-identification [7, 8].

The paradigm of deep metric learning centers on constructing of proper loss functions to constrain the distance between the anchor point and its positive and negative samples [9-11]. The common goal of these loss functions is to increase the distance between the anchor point and its negative samples while decrease the distance between the anchor point and its positive samples, thus maintaining an appropriate margin between the positive and negative samples. These methods rely on sample mining [12, 13], sample enhancement [14-16], and pair weighting strategies [17] to construct valid sample pairs and mine effective information between difficult sample pairs in a minibatch. Generally, these methods learn more discriminative embedding space by maximizing the variance between different classes. In this embedding space, samples of the same classes are compressed into individual clusters, while samples of different classes are far apart.

However, compressing similar samples together in a violent way completely ignores the variance between them during training. Traditional deep learning methods treat all positive samples equally due to the lack of sufficiently fine-grained sample labels, and focus attention on distinguishing negative samples. This can lead to unconscious disruption of the local structure among similar samples.

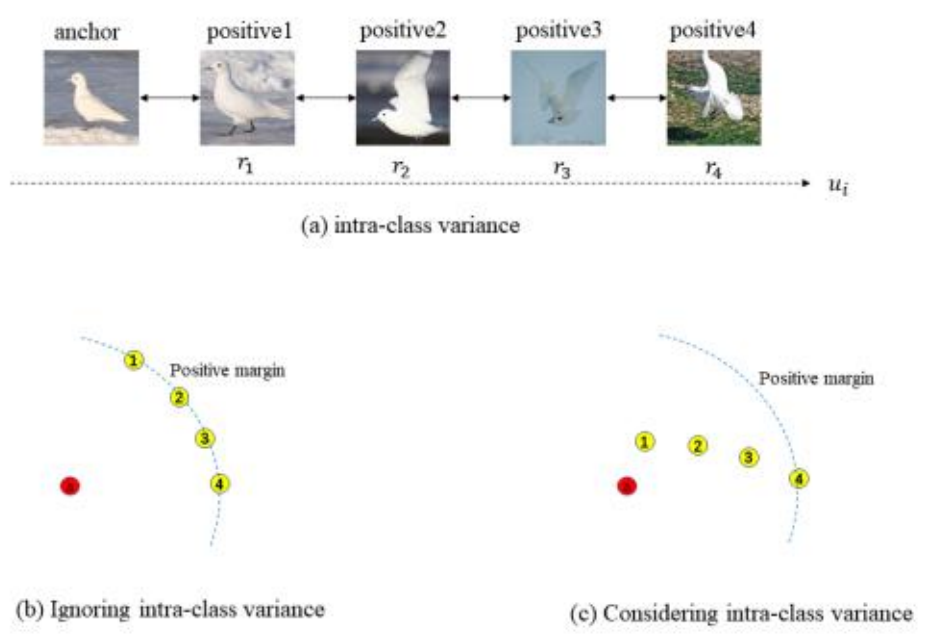

**Figure 1: Example of intra-class variance, and the effect on the embedding space. (a) $u_i$ is semantic direction (bird pose). $r$ is the strength of semantic change. The images with greater semantic strength (greater pose changes) are obviously less similar to the anchor. (b) Embedding space ignoring intra-class variance. (c) Embedding space considering intra-class variance.**

As shown in Figure 1(a), even similar samples have different intra-class variances. and previous deep metric learning methods ignore this property. Their embedding space is shown in Figure 1(b). Simply making the distance between similar positive samples and anchor less than a fixed margin does not reflect the difference between like samples, which decreases the quality of the embedding space. This reduces the generalizability of the model, making it overfit on the training set and perform poorly on the test set or on tasks where a large number of positive samples exist. Therefore, we designed a model that maximizes the inter-class variance while maintaining the intra-class variance. The distribution of samples in the embedding space learned by our method is shown in Figure 1(c). In this space, while the distance between positive samples and the anchor point is preserved less than the positive margin, the ranking relationship between positive samples is maintained according to their intra-class variance. For this consideration, we propose a self-supervised generative assisted ranking (SGAR) framework, which uses samples which satisfy the conditions to generate similar samples with measurable intra-class variance. Then we utilize the idea of self-supervised learning to design a ranking loss function to maintain the ranking relationships of the generated samples. It is worth noting that, in contrast typical deep metric learning methods which focus on inter-class variance, our method focuses on intra-class variance as a complement to inter-class variance. In contrast to self-supervised contrastive learning which use one sample within a class to construct augmented samples and compute loss function such as InfoNCE loss [18] to emphasize distinguishing between intra-class samples and inter-class samples, our method uses several intra-class samples to generate similar samples with measurable semantic variations and computes and designs a ranking loss to construct a ranking relationship between similar positive samples in the embedding space. In contrast to semi-supervised methods such as Fixmatch [19] which stress consistent output of intra-class samples across tasks, our method stresses the difference of intra-class samples. Figure 2 is the overall framework of our proposed method.

In conclusion, our method makes full use of the neglected information between positive samples to mine the intra-class variance through a self-supervised process, and protects the local structure of samples in the embedding space from being destroyed. The similar work as ours is [20]. The difference is that we design a dynamic sample selection strategy that applies the mining of intra-class variance dynamically during training. And in addition, we design a powerful ranking loss function to better constrain the ranking relationship. To verify the validity of our method, we perform a wide range of experiments on four commonly used fine-grained datasets, including CUB-200-2011 (CUB) [21], Cars-196 (CARS) [22], Stanford Online Products (SOP) [23] and In-shop Clothes Retrieval (In-Shop) [24].

The main contributions of our work are as follows:

(1) We make full use of the positive samples which are ignored in typical deep metric learning methods to generate similar samples whose intra-class variance is metrizable. Based on this, we use the idea of self-supervised learning to maintain the intra-class ranking relationship on the basis of the generated samples, thus mining the intra-class variance, protecting the local structure of the embedding space from being destroyed, and improving the quality of the embedding space.

(2) We design an efficient loss function that allows the model to better mine intra-class variance while also ensuring inter-class variance to learn a more discriminative embedding space.

(3) We conducted extensive experiments and made extensive comparisons with other methods to demonstrate the effectiveness of our proposed method.

The rest of this paper is divided into four sections. Section 2 presents related work about deep metric learning, self-supervised learning and sample generation. Section 3 describes our proposed method and loss function in detail. Section 4 is our experiment. Section 5 is the conclusions of our SGRA method.

# 2 RELATED WORK

## 2.1 Deep Metric Learning

Deep metric learning aims to learn an embedding space in which the similarity or distance between samples can be well measured, while samples of the same class are close to each other, and samples of the different class are far from each other. It is wildly used in the field of content-based image retrieval [2, 3], medical image processing [4-6], person re-identification [7, 8] and face recognition [25]. Deep metric learning mainly focuses on the design of loss functions, which are currently divided into two main categories: pair-based losses [26, 27] and proxy-based losses [28]. Pair-based losses construct sample pairs from the training batch, and design optimizing objective based on the pairs. Typical pair-based losses include Contrastive loss [26, 27], Triplet loss [11], Lifted structured loss [23], N-pair loss [29], Margin-based loss [30] and Multi-Similarity loss [10]. However, the training complexity of pair-based losses, due to the need to construct sample pairs, often reaches $O(N^2)$ or $O(N^3)$, where $N$ is the number of samples in the training set. Proxy-based losses [28] can reduce the high complexity of pair-based losses. The common idea of proxy-based losses is to assign one or more proxies to each class, and use sample-to-proxy interactions instead of sample-to-sample interactions to learn the global structure of embedding space. The training complexity of proxy-based losses is $O(MN)$, where $M$ is the number of classes and $N$ is the number of samples. Because the number of classes is much smaller than the number of samples for most datasets, the training complexity of proxy-based losses is much lower than that of pair-based losses. Typical proxy-based losses including Proxy-NCA loss [28] and Proxy Anchor loss [9].

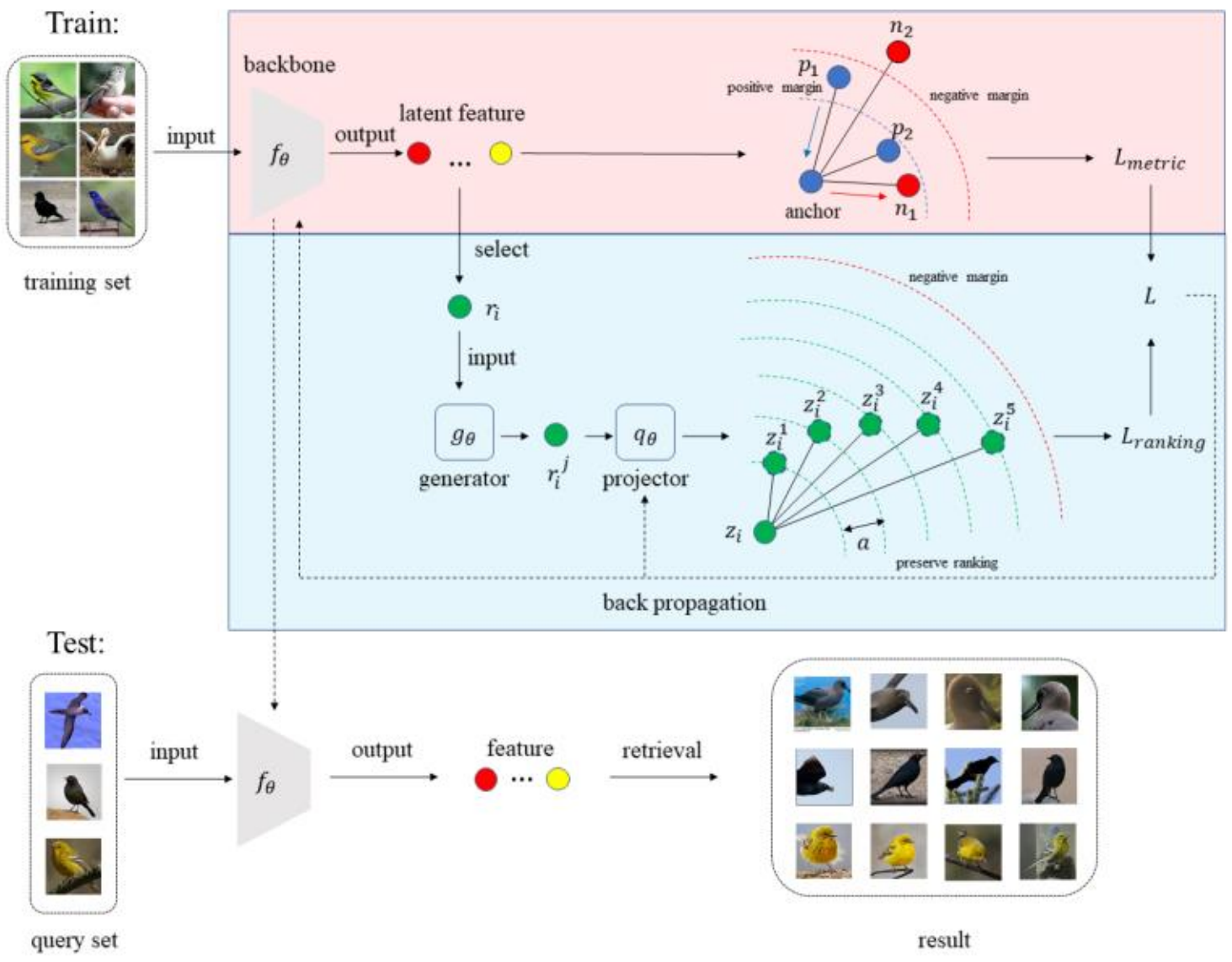

**Figure 2: The overall framework of our proposed approach. The part in red background is a typical deep metric learning framework and $L_{metric}$ is a deep metric learning loss. And the part in blue background is our self-supervised generative assisted ranking framework. Our framework selects appropriate features to generate similar samples $r_i^j$ from the latent embedding space $r_i$ obtained from the backbone $f_\theta$. Then we map these generated samples to a new embedding space $z_i^j$ via projector $q_\theta$ and use a powerful self-supervised ranking loss function $L_{ranking}$ to constrain the ranking relationship of similar samples. Eventually, $L_{metric}$ and $L_{ranking}$ are combined and optimize the network by backpropagation. In testing, we only use the trained backbone to extract features, calculate the similarity and return the retrieval results.**

## 2.2 Intra-class Variance

Intra-class variance refers to the difference between samples of the same class. Most researches focus on the differences between samples of different classes, which is the inter-class variance, and ignore the intra-class variance. However, some recent studies learn intra-class variance in different way and show that the intra-class variance contributes significantly to the quality of the embedding space. Wang et al. [31] designed a dynamic margin instead of the fixed margin used in typical deep metric learning loss functions, which dynamically selects the size of the margin according to the intra-class variance of different classes and better maintains the intra-class variance. Fu et al. [32] use different scales of image augmentation to model the intra-class variation of samples, and in their another work [20] devise a better sample generation approach to generate samples with measurable intra-class variance in the embedding space, and then maintaining the intra-class variance by constraining the ranking relationship between them.

# 3 METHOD

## 3.1 Intra-class Variance

Intra-class variance refers to the difference between samples of the same class. The opposite is inter-class variance, which means the difference between samples of different classes. We suppose that the image set is denoted as $X = \{x_1, \cdots, x_M\}$, and the corresponding label set is $Y = \{y_1, \cdots, y_M\}$, where $M$ denotes the size of the set. For a sample $x_{\mathrm{a}}$, we define the positive sample $x_p \in X_P$ and negative sample $x_n \in X_N$, while $y_{\mathrm{a}} = y_{\mathrm{p}} \neq y_{\mathrm{n}}$. After mapping $x_{\mathrm{a}}$, $x_p$ and $x_n$ to embedding space through a deep neural network, they are denoted as $z_{\mathrm{a}}$, $z_{\mathrm{p}}$ and $z_{\mathrm{n}}$. Then intra-class variance and the inter-class variance can be represented as $d(z_{\mathrm{a}}, z_{\mathrm{p}})$ and $d(z_{\mathrm{a}}, z_{\mathrm{n}})$. Typical deep metric learning methods use pulling close and pushing away positive and negative sample pairs to learn the embedding space, where the following constraint is guaranteed:

$$\max_{x_p \in X_p} d(z_a, z_p) < \min_{x_n \in X_n} d(z_a, z_n) \quad (1)$$

However, the constraint focuses on maximizing inter-class variance and insufficiently exploits intra-class variance. This unconsciously destroys the local structure between similar samples and waste the information contained in the intra-class variance. In order to use this information, we hope that samples with smaller intra-class variance in the original space will also be closer in the embedding space:

$$if\ d_{origin}(x_a, x_{p1}) < d_{origin}(x_a, x_{p2})$$
$$then\ d(z_a, z_{p1}) < d(z_a, z_{p2}) \quad (2)$$

Under this constraint, the local structure between similar samples is also well preserved, which is useful for distinguishing fine-grained datasets and increasing generalizability over unseen classes.

## 3.2 Sample Generation

The precondition for generating samples in embedding space is that the deep neural network can capture the semantic features of the image [33]. Several researches [34, 35] have shown that translating a feature along certain semantic directions in the embedding space produces features with meaningful different identities. Recently, some work [36, 37] has made promising progress by generating samples in embedding space instead of traditional data augmentation. Fu et al. [20] proposed a sample generation method with measurable semantic variance, which provides the possibility of capturing intra-class variance.

Specifically, input samples into a Convolutional Neural Network (CNN), we can get the features $R = \{r_1, \cdots, r_K\}$ in the latent space through a non-linear transformation $r_i = f(x_i, \theta_f)$, where $f(\cdot)$ denotes the embedding and the $\theta_f$ is the parameters of a CNN.

For sample $r_i$ and its generated sample $r_i^j$, we can decompose the relationship between them into scalar and direction in the polar coordinate system:

$$r_i^j = r_i + \alpha u_i^j \quad (3)$$

where $\alpha = \|r_i^j - r_i\|_2$ and $u_i^j = \frac{r_j - r_i}{\|r_j - r_i\|_2}$. In this way, we can use existing samples to synthesize similar samples by controlling the strength (scalar in polar coordinate) and semantic direction (direction in polar coordinate). The final generation formula is as follows:

$$r_i^j = r_i + j\alpha u_i^j \quad j = 1,2,\ldots,N \quad \|u_i^j\|_2 = 1 \quad (4)$$

where $j$ is a positive integer from 1 to $N$ which used to control the number of generated samples, $\alpha$ is a hyperparameter to ensure sample $r_i^N$ which generated using the maximum semantic variation $N\alpha$ is still similar to the original sample $r_i$, and $u_i^j$ is sampled from the standard Gaussian distribution.

As the intra-class variance space of $r_i$ is consistent of the generated local neighbor $r_i^j$ in latent space, the following constraint is satisfied:

$$if\ j_1 < j_2 \quad then\ d_l(r_i, r_i^{j_1}) < d_l(r_i, r_i^{j_2}) \quad (5)$$

where $d_l(r_i, r_i^j)$ denotes the distance between $r_i$ and $r_i^j$ in latent space. In this way, we can generate similar samples according to different semantic variations by controlling $j$, and the intra-class variance between them is measurable. As the samples are generated dynamically in embedding space during training, when the model is not well trained, the generated samples have a high error rate, which will instead affect the effectiveness of subsequent training. For this consideration, we propose a dynamic sample selection strategy which uses a generation margin $\gamma$ to control the selection of samples for generation, as shown in figure 3, which adaptively selects the samples used for generation with the quality of the embedding space.

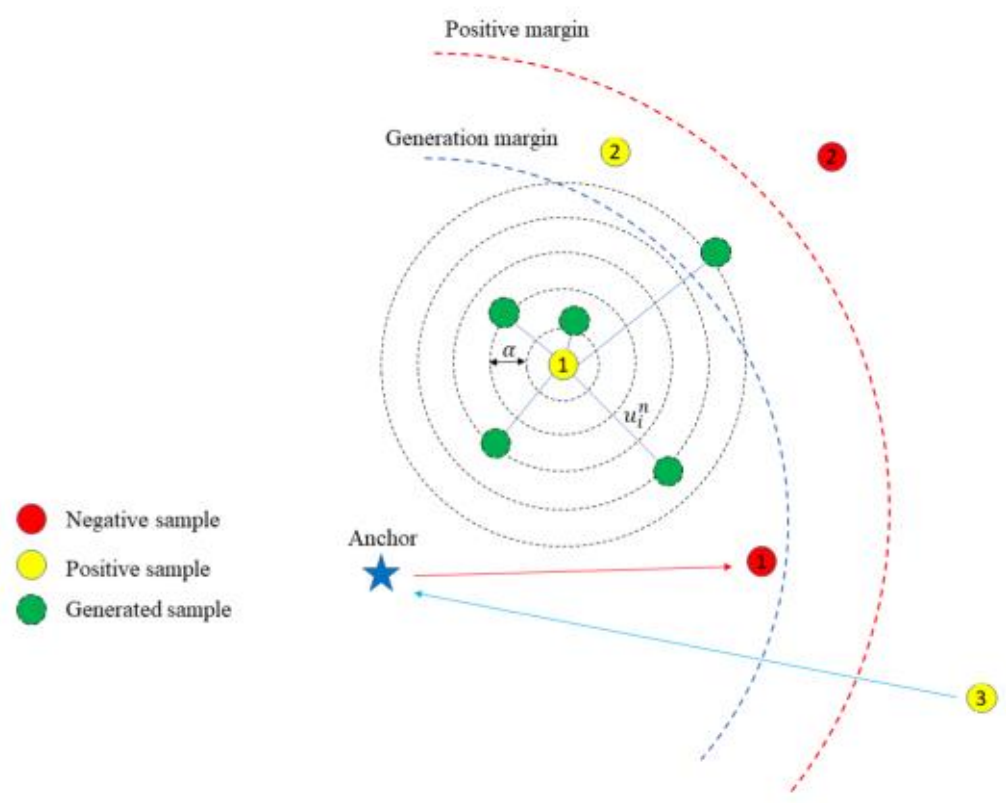

**Figure 3: Dynamic sample selection strategy. Only positive samples within generation margin (positive sample 1) are selected for sample generation. Negative sample 1 satisfies generation margin but not the positive margin and should be pushed away. Positive sample 2 is within positive margin and outside the generation margin, so it is not processed. Positive sample 3 is outside positive margin and should be pulled towards the anchor. Negative sample 3 is outside the positive margin and is not processed.**

In the early training period, the distance between most positive samples and anchor points is greater than the generation margin, and only a small number of positive samples are smaller than the generation margin. At this time, our generation strategy is very conservative, and generates few samples. When the network is well trained, more samples are smaller than the generation margin and are selected for generation. This strategy is motivated by an intuition that inter-class variance plays a major role in discrimination between samples of different classes, and using intra-class variance alone cannot distinguish samples of different classes well. However, focusing only on inter-class variance leads the model ignoring the intra-class distribution of samples and overfitting the training samples. Therefore, we want to use intra-class variance as an assist of inter-class variance to maintain intra-class distribution and improve the generalization of the model. Hence, in the preliminary of training, we want to focus on inter-class variance to initially construct a reasonable embedding space, and gradually increase the weight of intra-class variance as the quality of the embedding space improves to maintain the local distribution of similar samples.

With our designed generation strategy, the mining of intra-class variance is adaptively applied in training process. In addition, our strategy can improve the accuracy of the generated samples and avoid the slow training due to all or random selection of positive samples for generation.

## 3.3 Self-supervised Ranking Preserving

Using the sample generation strategy, we obtain the original sample $r_i$ and its generated samples $r_i^j$ in the latent space, where $r_i^j \in \{r_i^1,\cdots,r_i^N\}$ and $N$ is the number of samples generated per positive sample. Afterwards, we apply the idea of self-supervised learning, using a projector $q(\cdot)$ to map the latent space into a new embedding space, where $z_i = q(r_i, \theta_q)$ and $z_i^j = q(r_i^j, \theta_q)$. And we design an efficient loss function to learn the local ranking structure. For a uniform representation, in the later we use the cosine similarity instead of the distance above, since all embedding space are L2-normalized.

Our loss function is based on a ranking loss function as Eq. (6):

$$L_{hand-in-hand} = \frac{1}{M}\sum_{x=1}^{M}\sum_{i=1}^{N-1}\left[-S_{i,x} + S_{i+1,x} + \delta\right]_+ \quad (6)$$

where $S_{i,x}$ denotes the similarity of the $i$-th generated sample to the original sample, $\delta$ is the margin between two adjacent generated samples. By compute this loss function, the similarity between every two adjacent generated samples differs by a fixed margin $\delta$, and the smaller numbered generated samples (samples with lower intra-class variance strength) are more similar to the original samples. However, in this loss function, for a generated sample, only the two samples before and after it are considered. We therefore split it into left part and right part, and add consideration of the global ranking of the generated samples as Eq. (7) and Eq. (8):

$$L_{left_base} = \frac{1}{M}\sum_{x=1}^{M}\sum_{i=2}^{N}\left[S_{i,x} - \min_{j\in[1,i-1]} S_{j,x} + \delta\right]_+ \quad (7)$$

$$L_{right_base} = \frac{1}{M}\sum_{x=1}^{M}\sum_{i=1}^{N-1}\left[\max_{j\in[i+1,N]} S_{j,x} - S_{i,x} + \delta\right]_+ \quad (8)$$

For a generated sample, Eq. (7) constrains the least similar one (has the lowest similarity) among all samples to its left (smaller numbering) to be ranked ahead of it. While Eq. (8) constrains the sample with the greatest similarity on the right side to be ranked after the current sample. Figure 4 illustrates how our loss works:

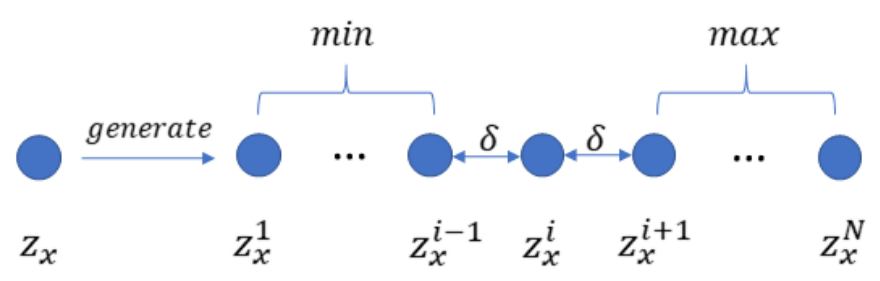

**Figure 4: Explanation of Eq. (7) and Eq. (8). $z_x$ is original sample. $z_x^1$ to $z_x^N$ are the samples generated using $z_x$ as the prototype. For $z_x^i$, Eq. (7) constraints the sample with the least similarity on its left side to be greater than its similarity by a margin $\delta$. And Eq. (8) constraints the sample with the largest similarity on its right side to be less than its similarity by a margin $\delta$.**

we apply the LogSumExp [10] and SoftPlus [38] functions, and finally get a smooth version of our function. The loss becomes:

$$L_{left} = \frac{1}{M}\sum_{x=1}^{M}\frac{1}{\tau}\log\left(1 + \sum_{i=2}^{N}\sum_{j=1}^{i-1} e^{\tau(S_{i,x}-S_{j,x}+\delta)}\right) \quad (9)$$

$$L_{right} = \frac{1}{M}\sum_{x=1}^{M}\frac{1}{\tau}\log\left(1 + \sum_{i=1}^{N-1}\sum_{j=i+1}^{N} e^{\tau(S_{j,x}-S_{i,x}+\delta)}\right) \quad (10)$$

where $\tau$ is scaling parameter. Our sorting loss is combined by $L_{left}$ and $L_{right}$ as Eq. (11):

$$L_{sort} = L_{left} + L_{right} \quad (11)$$

The problem of Eq. (7) and Eq. (8) is that they have a fixed gradient which values of 1 or -1. This approach lacks the ability to mine informative sample pairs [10] and leads to trivial samples [39]. In contrast, the derivatives of Eq. (9) and Eq. (10) assign weights more dynamically to the degree of violation of the constraint by sample pairs. And compare to Eq. (6), Eq. (9) and Eq. (10) comprehensively consider the effect of all samples before and after the current sample. As shown in Eq. (12) and Eq. (13):

$$\frac{\partial L_{left}}{\partial S_{i,x}} = \frac{\sum_{j=1}^{i-1} e^{\tau(S_{i,x}-S_{j,x}+\delta)}}{1+\sum_{i=2}^{N}\sum_{j=1}^{i-1} e^{S_{i,x}-S_{j,x}+\delta}} \quad (12)$$

$$\frac{\partial L_{right}}{\partial S_{i,x}} = -\frac{\sum_{j=i+1}^{N} e^{\tau(S_{j,x}-S_{i,x}+\delta)}}{1+\sum_{i=1}^{N-1}\sum_{j=i+1}^{N} e^{S_{j,x}-S_{i,x}+\delta}} \quad (13)$$

Although our method produces generated samples through a series of semantic transformations, these samples are of the same class as the original sample and need to maintain the constraint with the anchor:

$$L_{anchor} = \frac{1}{|P^+|}\sum_{p\in P^+}\frac{1}{\beta}\sum_{x\in X_P}\left(\log\left(1 + \sum_{i=1}^{N} e^{\beta(\varphi-S_{i,p})}\right)\right) \quad (14)$$

where $P^+$ is the set of positive anchors, $p \in P^+$ is the positive anchor, $\beta$ is the scaling parameter, $\varphi$ is a margin and $S_{i,p}$ denotes the similarity of generated samples and the positive anchor. The derivate of Eq. (14) is shown as Eq. (15), where the pairs with lower similarity have a larger weight.

$$\frac{\partial L_{anchor}}{\partial S_{i,p}} = -\frac{e^{\beta(\varphi-S_{i,p})}}{1+\sum_{i=1}^{N} e^{\beta(\varphi-S_{i,p})}} \quad (15)$$

Combining Eq. (11) and Eq. (14), we finally propose the ranking loss as:

$$L_{ranking} = L_{sort} + L_{anchor} \tag{16}$$

Our method can be viewed as an independent module and can be plug-and-play combined with typical metric learning methods. We balance our ranking loss $L_{ranking}$ and metric learning loss $L_{metric}$ with a hyperparameter $\lambda$.

$$L = L_{metric} + \lambda L_{ranking} \tag{17}$$

# 4 EXPERIMENTS

## 4.1 Dataset

We use four widely-used dataset to evaluate our proposed method. They are CUB-200-2011 (CUB) [21], Cars-196 (CARS) [22], Stanford Online Product (SOP) [23] and In-shop Clothes Retrieval (InShop) [24]. CUB is a bird dataset containing 200 different classes with 11,788 images. For CUB, we use 5864 images from the first 100 classes for training, and 5924 images from the other 100 classes for testing. CARS is a car dataset containing 196 different car models with 16185 images. For CARS, we use 8,054 images of the first 98 classes for training, and the other for testing. SOP contains 120,053 images of 22,634 classes, which is online product sold on eBay.com. For SOP, we use the same split as [23], 59,551 images of 11,318 classes is used for training, and the other is used for testing. InShop is a clothes dataset, which contains 72,712 images of 7986 classes. For InShop, we use the same split as [24], which use 25,882 images of the first 3,997 classes for training and the other for testing. The testing set is further divided into query set and gallery set. The query set contains 14,218 images of 3,985 classes and the gallery set contains 12,612 images of 3,985 classes.

## 4.2 Baselines and Evaluation Metrics

We selected a large number of methods to compare and fully evaluate our method. Specifically, the methods used for comparison include some robust deep metric learning methods such as A-BIER [40], ABE [41], HTL [42] and RLL-H [39], some pair-based loss functions such as Circle loss [25] and Multi-Similarity loss [10] and proxy-based loss such as SoftTriple loss [43], Proxy-GML loss [44] and Proxy-Anchor loss [9]. We also chose some sample generation methods that are similar to our approach including SR [32] and SSR [20]. To ensure comprehensive evaluation, we use various evaluation metrics. In order to accommodate most of methods, we choose $Recall@K$ ($R@K$), which is commonly used in retrieval, as the evaluation metric.

## 4.3 Implementation Details

*4.3.1* Embedding Networks

In order to test the effectiveness of our method on different networks and to compare fairly with previous works, we use GoogleNet [45] and BN–Inception with batch normalization [46] as the backbone network respectively. The backbone is pre-trained for ImageNet classification [47]. The projector $q_\theta$ is composed of a fully connected layer, an L2-normalization layer and a Relu activation function layer. The dimension of $q_\theta$ is 512.

*4.3.2* Training

We use random crop and random horizontal flip for data augmentation in training and only single center crop in testing. The size of cropped images is set to $224 \times 224$ as in most previous researches. The AdamW optimizer [48] is used in every experiment. For CUB and CARS, we use $10^{-4}$ as initial learning rate, and train for 40 epochs. For SOP and InShop, the initial learning rate is set to $6 * 10^{-4}$ and train for 60 epochs.

*4.3.3* Hyperparameters

We set $\delta$ and $\tau$ to 0.3 and 64 in Eq. (9) and Eq. (10). $\alpha$ and $N$ in Eq. (4) is set to 1.0 and 5. Generation margin $\gamma$ is set to 0.05. $\lambda$ in Eq. (17) is set to 0.1. $\varphi$ in Eq. (14) is set to 0.1. For the batch size, we use the settings as [9]. Specifically, we set the batch size to 180 on CUB and CARS and 120 on SOP and InShop.

## 4.4 Experimental Results and Analysis

We evaluate our method based on Proxy-Anchor loss [9] and using the same settings as in their paper. We select some excellent sample generation-based approaches and some landmark works in deep metric learning to compare with our method. For fairness, we use the same embedding size and backbone. Figure 5 shows our qualitative results:

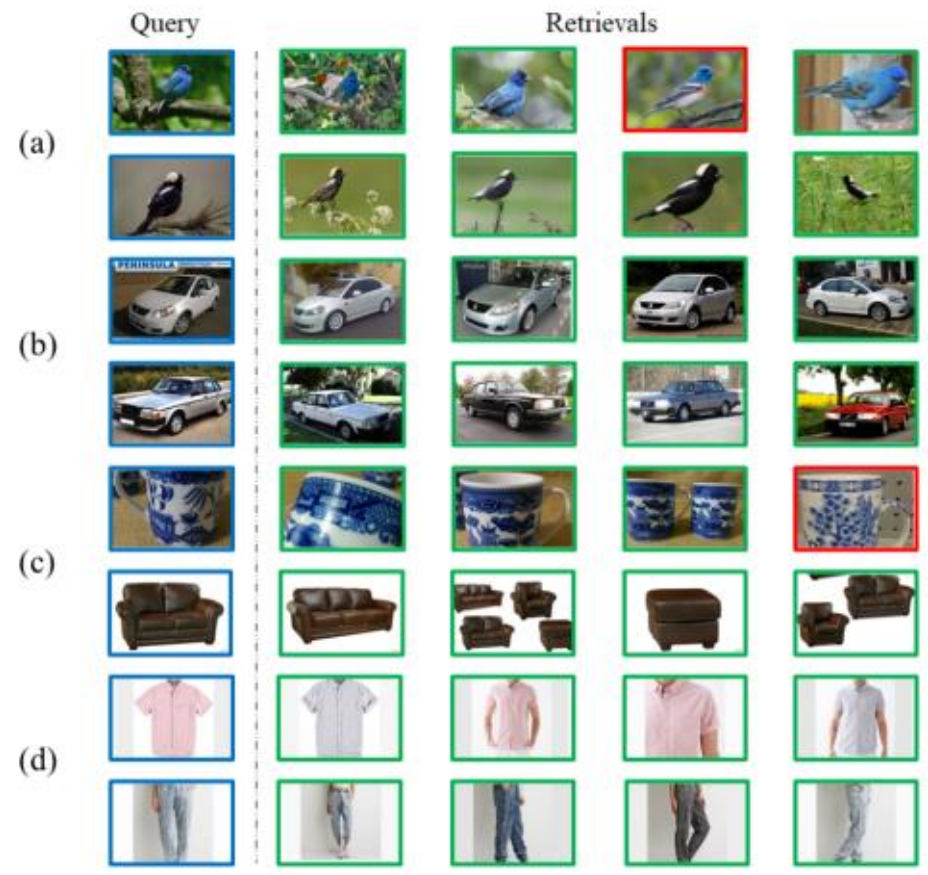

**Figure 5: Qualitative retrieval results of our method. (a), (b), (c), (d) is the top-5 recall results on CUB, CARS, SOP and Inshop. The left side of the dotted line with blue edge indicates query images. The right side of the dotted line shows the retrieval results, where the green edges indicate correct results and the red edges indicate incorrect results.**

A comparison of our approach with the state-of-art deep metric learning methods is presented and the results are shown in Table 1, Table 2 and Table 3. Since the different backbone and embedding size greatly affect performances, we list the settings used by different methods in tables.

On CUB, for excellent deep metric learning methods such as A-BIER, ABE, HTL and RLL-H, we use GoogleNet as backbone and achieve improvements of 6.8%, 3.7%, 7.2% and 6.9% in

in $R@1$ than SR and SSR by 57.4% → 64.3% and 65.4% → 68.8

$R@1$. For pair-based losses such as Circle and MS, we use BN-Inception as backbone and achieve improvements of 2.1% and 3.1% in $R@1$. For proxy-based losses such as SoftTriple, Proxy-GML and Proxy-Anchor, we obtain performance boost in $R@1$ for 3.4%, 2.2% and 0.4%. We still achieve a higher performance

**Table 1: Comparison with excellent method on CUB and CARS. Method indicates the different methods. Setting indicates the backbone used by current method, where G denotes GoogleNet [45], BN denotes BN-Inception with batch normalization [46] and RN denotes ResNet50 [49]. The number in the superscript in the setting indicates embedding size. Recall@K is used as evaluation metric. R denotes recall and the number after it denote K. Since Proxy-Anchor loss [9] does not report results on GoogleNet, we reproduce the results and mark the results with the superscript *.**

| | | CUB-200-2011 | | | | Cars-196 | | | |
|---|---|---|---|---|---|---|---|---|---|
| Method | setting | R1 | R2 | R4 | R8 | R1 | R2 | R4 | R8 |
| A-BIER [40] | $G^{512}$ | 57.5 | 68.7 | 78.3 | 86.2 | 82.0 | 89.0 | 93.2 | 96.1 |
| ABE [41] | $G^{512}$ | 60.6 | 71.5 | 79.8 | 87.4 | 85.2 | 90.5 | 94.0 | 96.1 |
| HTL [42] | $BN^{512}$ | 57.1 | 68.8 | 78.7 | 86.5 | 81.4 | 88.0 | 92.7 | 95.7 |
| RLL-H [39] | $G^{512}$ | 57.4 | 69.7 | 79.2 | 86.9 | 74.0 | 83.6 | 90.1 | 94.1 |
| SoftTriple [43] | $BN^{512}$ | 65.4 | 76.4 | 84.5 | 90.4 | 84.5 | 90.7 | 94.5 | 96.9 |
| Proxy-GML [44] | $BN^{512}$ | 66.6 | 77.6 | 86.4 | - | 85.5 | 91.8 | **95.3** | - |
| Circle [25] | $BN^{512}$ | 66.7 | 77.4 | 86.2 | 91.2 | 83.4 | 89.8 | 94.1 | 96.5 |
| MS [10] | $BN^{512}$ | 65.7 | 77.0 | 86.3 | 91.2 | 84.1 | 90.4 | 94.0 | 96.5 |
| SR [32] | $G^{512}$ | 57.4 | 69.3 | 79.8 | - | 80.9 | 88.2 | 92.6 | - |
| SSR [20] | $BN^{512}$ | 65.4 | 76.3 | 84.3 | 80.3 | 80.1 | 87.3 | 92.2 | 95.1 |
| Proxy-Anchor* | $G^{512}$ | 63.9 | 74.8 | 83.7 | 89.9 | 84.7 | 90.6 | 94.5 | 96.9 |
| Proxy-Anchor+SGAR | $G^{512}$ | 64.3 | 75.1 | 83.9 | 90.4 | 85.7 | 91.0 | 94.8 | 97.1 |
| Proxy-Anchor [9] | $BN^{512}$ | 68.4 | 79.2 | 86.8 | 91.6 | 86.1 | 91.7 | 95.0 | 97.3 |
| Proxy-Anchor+SGAR | $BN^{512}$ | **68.8** | **79.4** | **86.9** | **91.7** | **86.6** | **92.0** | 95.1 | **97.4** |

On CARS, we use the same method as on CUB for comparison. For A-BIER, ABE, HTL and RLL-H, we obtain higher $R@1$ with boost of 3.7%, 0.5%, 4.3% and 11.7%. For Circle loss and MS loss, our $R@1$ rises by 3.2% and 2.5%. For SoftTriple, Proxy-GML and Proxy-Anchor, we boosted 2.1%, 1.1% and 0.5% on them in $R@1$. And we exceed SR and SSR by 4.8% and 6.5% in $R@1$.

On SOP, we still achieve different degrees of improvement compare to other methods. Our results outperform HTL, SoftTriple, Proxy-GML, Circle, MS, SR, SSR and Proxy-Anchor by 4.6%, 1.1%, 1.4%, 1.1%, 1.2%, 0.8%, 0.5% and 0.8% in $R@1$.

On InShop, our method is presented in 11.0%, 2.2%, 0.9% and 0.3% ahead of HTL, MS, SSR and Proxy-Anchor.

Experiments show that our method achieves the state-of-the-art results on four benchmark datasets. Our method captures intra-class variance by constructing intra-class ranking model to improve the stability and generalization of the embedding space. In the embedding space learned by our method, not only samples of different classes are separated, but also ranking relationships are maintained between similar samples. In contrast, other methods are misled by the presence of various intra-class variance in similar samples such as poses, background or viewpoints, which leads to worse results. On SOP, our method achieves the greatest improvement on the basis of the results of Proxy-Anchor. This may be motivated by an observation: SOP is a dataset of artificial products with a large perspective transformation between similar samples, while compare to CUB, SOP contains fewer samples in each class. Therefore, our method plays a more significant auxiliary effect. It is worth noting that although our method uses BN-Inception as backbone, on SOP it even outperforms SR and SSR with ResNet50 as backbone and our method also performs well on GoogleNet. Adding our auxiliary framework to networks with poorer representation can help the network learn more useful knowledge. This indicates the effectiveness of the intra-class information mined by our method and the prospects of our method for lightweight models.

## 4.5 Ablation Study

In order to verify the effectiveness of our method, we perform extensive ablation experiments. Specifically, we use Proxy-Anchor loss and train on CUB. BN-Inception with 512 embedding size is the default backbone.

Figure 6(a) shows the effect of the hyperparameter $\lambda$, which is a factor used to balance the benchmark deep metric learning loss and our ranking loss, and also reflects the proportion of inter-class variance and intra-class variance during training. When $\lambda$ is small, the intra-class variance has little impact in training and is not sufficiently powerful to optimize the embedding space. And when $\lambda$ is large, the impact of intra-class variance exceeds inter-class variance, which leads to severe performance degradation. This is in line with our previous consideration that the inter-class variance plays a major role in distinguishing different classes of samples, but introduction of appropriate intra-class variance helps to maintain the local structure of the embedding space and improve generalization. The performance is best when $\lambda = 0.1$, so we set $\lambda = 0.1$.

**Table 2: Comparison with excellent method on SOP.**

| | | SOP | | | |
|---|---|---|---|---|---|
| Method | setting | R1 | R10 | R100 | R1000 |
| A-BIER [40] | $G^{512}$ | 74.2 | 86.9 | 94.0 | 97.8 |
| ABE [41] | $G^{512}$ | 76.3 | 88.4 | 94.8 | 98.2 |
| HTL [42] | $BN^{512}$ | 74.8 | 88.3 | 94.8 | 98.4 |
| RLL-H [39] | $G^{512}$ | 76.1 | 89.1 | 95.4 | - |
| SoftTriple [43] | $BN^{512}$ | 78.3 | 90.3 | 95.9 | - |
| Proxy-GML [44] | $BN^{512}$ | 78.0 | 90.6 | 96.2 | - |
| Circle [25] | $BN^{512}$ | 78.3 | 90.5 | 96.1 | - |
| MS [10] | $BN^{512}$ | 78.2 | 90.5 | 96.0 | 98.7 |
| SR [32] | $RN^{512}$ | 78.6 | 90.6 | 96.2 | 98.7 |
| SSR [20] | $RN^{512}$ | 78.9 | 91.0 | 96.2 | 98.8 |
| Proxy-Anchor [9] | $BN^{512}$ | 78.6 | 90.3 | 95.9 | 98.6 |
| Proxy-Anchor+SGAR | $BN^{512}$ | 79.4 | 91.0 | 96.4 | 98.8 |

**Table 3: Comparison with excellent method on Inshop.**

| | | InShop | | | |
|---|---|---|---|---|---|
| Method | setting | R1 | R10 | R20 | R40 |
| A-BIER [40] | $G^{512}$ | 83.1 | 95.1 | 96.9 | 97.8 |
| ABE [41] | $G^{512}$ | 87.3 | 96.7 | 97.9 | 98.5 |
| HTL [42] | $BN^{512}$ | 80.9 | 94.3 | 95.8 | 97.4 |
| MS [10] | $BN^{512}$ | 89.7 | 97.9 | 98.5 | 99.1 |
| SR [32] | $RN^{128}$ | 88.0 | 97.3 | 98.2 | - |
| SSR [20] | $BN^{512}$ | 91.0 | 98.0 | 98.7 | - |
| Proxy-Anchor [9] | $BN^{512}$ | 91.6 | 98.1 | 98.8 | **99.1** |
| Proxy-Anchor+SGAR | $BN^{512}$ | **91.9** | **98.3** | **98.9** | **99.1** |

Figure 6(b) shows the results for different generation margin γ. The best result is obtained when γ=0.05, so we set γ=0.05. Only samples with a similarity greater than the generation margin γ are selected for generation and calculate the ranking loss, thus γ has a significant impact in training. A small γ indicates that the more samples used for generation, and more impact the intra-class variance has in training. But excessively decreasing γ will result in low quality of generated samples, which in turn misleads the training of the network. Using a large γ means that few samples in the training satisfy the generation condition, which reduces the performance of the intra-class variance.

Figure 6(c) and Figure 6(d) shows the results of α and δ. α and δ control the generation strength and the margin to be maintained between neighboring samples. According to the experimental results we set them respectively as 1.0 and 0.3. Figure 6(e) shows the results of τ and we set is to 64.

Figure 6(f) shows our result on N, where N represents the number of samples generated for each sample. Since our loss needs to consider the left and right parts, N is at least 3. A larger N means a larger number of generated samples, which leads to more time complexity for training, and a larger scale of variation in generated samples, which is difficult to control. A smaller N means a smaller number of generated samples and a more conservative use of intra-class variance. On balance, we set N to 5.

## 5 CONCLUSION

In this paper, we present a novel self-supervised generative assisted ranking framework that provides a semi-supervised perspective on the typical supervised deep metric learning methods as a boosting scheme. Our method does not require additional labels, generates samples using existing positive samples. Inspired by the idea of self-supervised learning, we design a powerful ranking loss function to preserve the ranking relationship among similar samples in the local space of the embedding space. Extensive experiments demonstrate that our approach can combine with and improve the performance of typical supervised deep metric learning methods and outperforms state-of-the-art methods on four benchmark datasets.

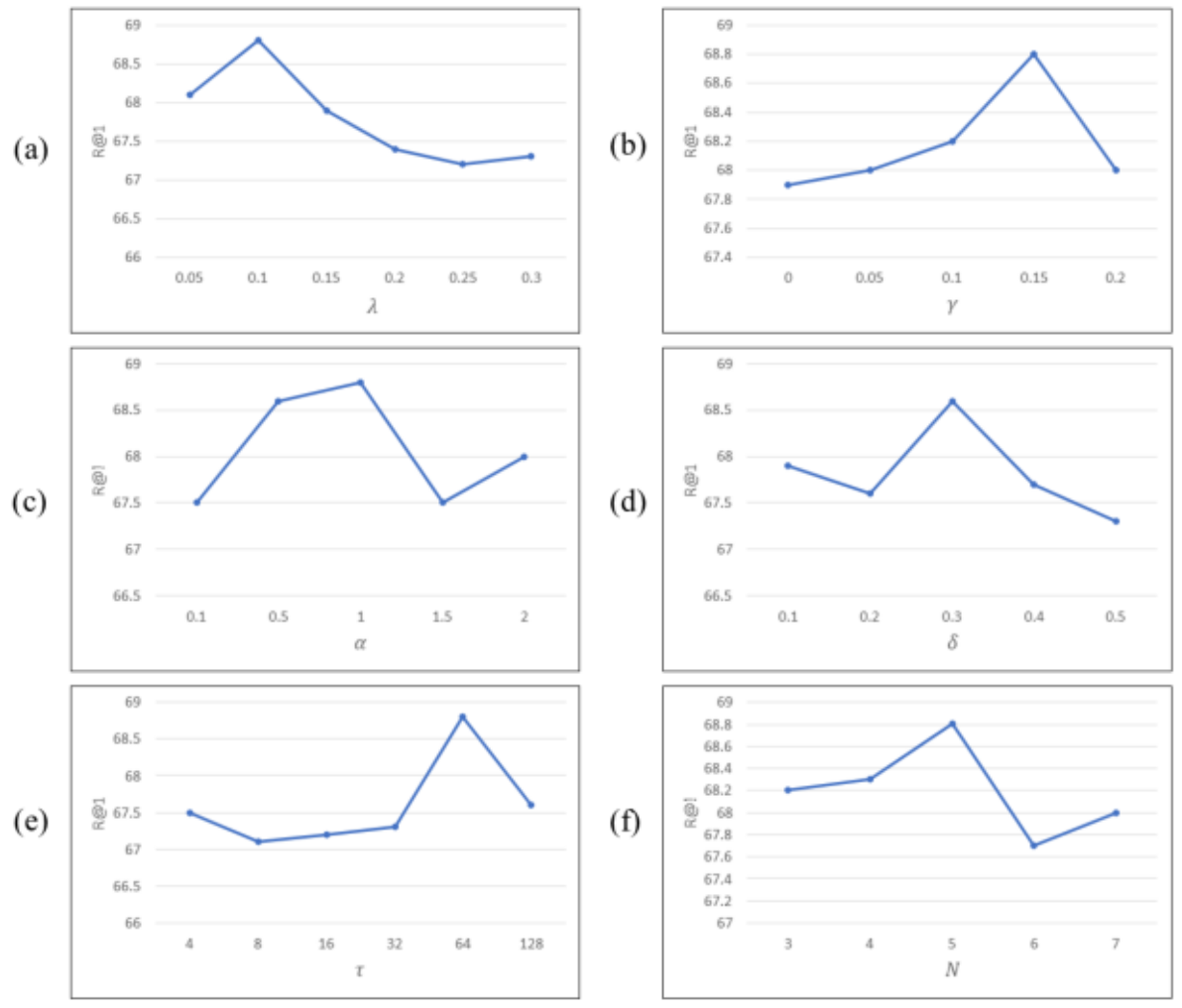

**Figure 6: R@1 with different settings of six significant hyperparameters. (a) factor $\lambda$ used to balance $L_{metric}$ and $L_{ranking}$, (b) generation margin $\gamma$, (c) generate strength scalar $\alpha$, (d) margin between neighbors $\delta$, (e) scale factor $\tau$, and (f) number of generated samples of each sample $N$.**

## ACKNOWLEDGMENTS

Insert paragraph text here. Insert paragraph text here. Insert paragraph text here. Insert paragraph text here. Insert paragraph text here. Insert paragraph text here. Insert paragraph text here. Insert paragraph text here. Insert paragraph text here. Insert paragraph text here. Insert paragraph text here. Insert paragraph text here.